%% file: arxiv.tex
\pdfoutput=1 
\documentclass{article}
\usepackage{iclr2027_conference,times}
\iclrfinalcopy 

\usepackage{amsmath,amssymb}
\usepackage{booktabs,tabularx}
\usepackage{placeins}
\usepackage{xspace}
\usepackage{graphicx}
\usepackage{caption}
\usepackage{tikz}
\usetikzlibrary{arrows.meta}
\usepackage{hyperref}
\usepackage{url}
\definecolor{citeblue}{RGB}{20,60,180}
\definecolor{linkred}{RGB}{180,30,30}
\hypersetup{colorlinks=true, citecolor=citeblue, linkcolor=linkred, urlcolor=citeblue}

\input{defs}

\makeatletter
\renewcommand{\paragraph}{\@startsection{paragraph}{4}{\z@}%
  {0.6ex plus 0.2ex minus 0.1ex}{-1em}{\normalsize\bfseries}}
\renewcommand{\subsection}{\@startsection{subsection}{2}{\z@}%
  {-1.0ex plus -0.3ex minus -0.2ex}{0.4ex plus 0.1ex}{\normalsize\sc\raggedright}}
\makeatother

\newcommand{\method}{StraTune\xspace}
\newcommand{\train}{\mathcal{D}_{\mathrm{train}}}
\newcommand{\test}{\mathcal{D}_{\mathrm{test}}}

\newcommand{\arxivheader}{Preprint} 
\hypersetup{pdftitle={StraTune: Adaptive Selection of Revision Operators for Self-Evolving LLM Skills},
            pdfauthor={Zeping Liu, Yan Li, Ni Lao, Gil Wolff, Gengchen Mai}}

\title{\method: Adaptive Selection of Revision Operators for Self-Evolving LLM Skills}

\author{Zeping Liu$^{1}$, Yan Li$^{2}$, Ni Lao$^{1}$, Gil Wolff$^{2}$ and Gengchen Mai$^{1,\dagger}$ \\[3pt]
{\normalfont\small $^{1}$The University of Texas at Austin, $^{2}$Amazon} \\[1pt]
{\normalfont\small $^{\dagger}$Corresponding author.}
}

\begin{document}
\setlength{\parskip}{.5pc}
\maketitle
\lhead{\arxivheader} 
{\renewcommand{\thefootnote}{}\footnotetext{\emph{Corresponding author:} \texttt{gengchen.mai@austin.utexas.edu}.}}
\suppressfloats[t]

\begin{abstract}
\input{sections/abstract}
Code and learned skills are available at \url{https://github.com/seai-lab/StraTune}.
\end{abstract}
\setlength{\parskip}{1pt plus 0.3pt minus 0.3pt}

\input{sections/introduction}
\input{sections/problem_statement}
\input{sections/method}
\FloatBarrier
\input{sections/experiments}
\input{sections/conclusion}

\section*{Reproducibility Statement}
The code of \method, the learned skills, and the data splits and processing scripts used in the experiments are available at \url{https://github.com/seai-lab/StraTune}. The method is specified in Section~\ref{sec:method}, with the search strategies, revision forms, and selection prompt in Appendix~\ref{app:search_strategies}--\ref{app:adaptive_selection} and the candidate evaluation rules in Appendix~\ref{app:evaluation_rules}. Appendix~\ref{app:experimental_details} lists the datasets and splits, model versions and decoding settings, the training budget, the baseline configurations, and the \method settings. All five baselines are run with their publicly released code, with the settings given in Appendix~\ref{app:baseline_settings}.

\bibliography{iclr2027_conference}
\bibliographystyle{iclr2027_conference}

\clearpage
\appendix
\input{sections/appendix}

\end{document}

%% file: defs.tex
\newcommand{\batch}{\mathcal{Q}}
\newcommand{\numbatch}{N_\batch}

%% file: sections/abstract.tex
Large language models (LLMs) can learn reusable textual skills from execution feedback without updating their parameters, but effectively deciding how to revise these skills remains a key challenge. Existing methods typically rely on a fixed revision operator, a search strategy and the revision forms applied under it. However, we observe that no single revision operator consistently performs best across tasks, and repeatedly applying an unsuitable operator can limit further improvement. We propose \textbf{StraTune} (\textbf{stra}tegy-guided skill \textbf{tun}ing), which lets a frozen optimizer LLM choose the revision operator at every round from the optimization state, which is defined as the current execution feedback together with the recorded outcomes of earlier strategies and forms. Candidate skills from every revision operator pass one candidate evaluation, which screens for gains and regressions on a small sample set and validates them on a larger one, and every outcome is written back to the optimization state for later choices. 
Across four benchmarks and two LLM settings, StraTune outperforms all five baselines in most settings. Ablations attribute the gains to the adaptive choice of the revision operator, since fixed, random, scheduled, and bandit strategy choices all score lower, and skills learned with a small target LLM also improve a stronger one.

%% file: sections/introduction.tex
\section{Introduction}
\label{sec:introduction}

\textbf{Large language models (LLMs)} increasingly rely on \emph{skills}, reusable text that tells a model how to perform a task, such as instructions, worked examples, and procedures, applied without changing the model's weights~\citep{brown2020language,wei2022chain,kojima2022large,liu2026spatial}. Skills have become a standard component of agent systems. An agent may load folders of instructions and scripts on demand~\citep{anthropic2025agentskills}, call programs from a skill library~\citep{wang2023voyager}, follow workflows induced for web navigation~\citep{wang2024agent,zheng2025skillweaver}, consult a memory of insights that carries across reasoning problems~\citep{suzgun2026dynamic}, read a playbook of strategies for agent and finance tasks~\citep{zhang2026agentic}, or carry domain knowledge into specialized fields~\citep{mai2025towards}.
Skills were first written by hand: a prompt engineer encodes expert task knowledge into instructions and tunes them by trial and error. Because this effort must be repeated for every new task and model, it motivated automatic search over the instruction text~\citep{zhou2022large,pryzant2023automatic,yang2024large,fernando2023promptbreeder,guo2024connecting,wang2024promptagent}. 
More recently, \textbf{self-evolving skills} have become common. Here, the \emph{target LLM}--the LLM that executes the task--receives execution feedback, and an \emph{optimizer LLM} turns the feedback into revisions of the skill~\citep{gao2025survey}, whether as verbal reflections~\citep{shinn2023reflexion,madaan2023self}, insights distilled from trajectories~\citep{zhao2024expel,ouyang2026reasoningbank}, code verified in the environment~\citep{wang2023voyager}, or textual gradients and edits~\citep{yuksekgonul2024textgrad,agrawal2026gepa,yang2026skillopt,ni2026trace2skill}.

\begin{figure}[t]
    \centering
    \includegraphics[width=\linewidth]{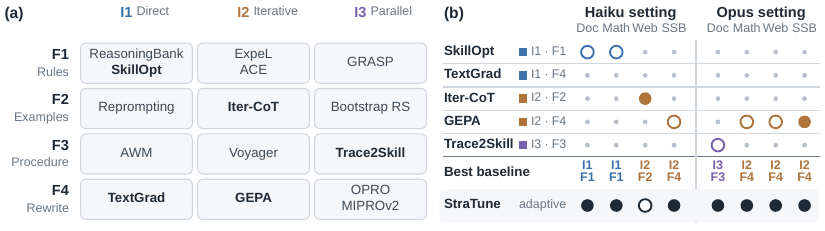}
    \caption{\textbf{(a)} Existing skill-learning methods by search strategy (columns) and revision form (rows); bold marks the baselines used in this paper. \textbf{(b)} Test scores (Table~\ref{tab:main_results}) on DocVQA (Doc), LiveMath (Math), Mind2Web (Web), and SpreadsheetBench (SSB) under two model settings; filled circles mark the best method in a column, open circles the second best. The row labelled \emph{Best baseline} gives the strategy and form of the best baseline in each column, which change from column to column, whereas \method ranks first in most.}
    \label{fig:design_space}
\end{figure}

Despite their variety, these methods share one loop. The target LLM rolls out the current skill on training tasks, a supervision signal (e.g., human judgments, an LLM verdict, or test
cases) marks each outcome as success or failure, and the optimizer LLM revises the skill. They differ in two choices made in every optimization round, and Figure~\ref{fig:design_space}(a) compares these choices and lists the corresponding existing methods. The first is \textbf{how candidate search is organized} (columns in Figure~\ref{fig:design_space}(a)). TextGrad~\citep{yuksekgonul2024textgrad} and SkillOpt~\citep{yang2026skillopt} revise the current skill directly (I1); GEPA~\citep{agrawal2026gepa} and Voyager~\citep{wang2023voyager} refine a separate
working version before it replaces the current skill (I2); OPRO~\citep{yang2024large}, MIPROv2~\citep{opsahl2024optimizing}, and Trace2Skill~\citep{ni2026trace2skill} generate several candidate skills and rank or merge them (I3). The second is \textbf{what kind of change is written into the skill} (rows in Figure~\ref{fig:design_space}(a)). SkillOpt, ExpeL~\citep{zhao2024expel}, and GRASP~\citep{moll2026grasp} add conditional rules (F1); Iter-CoT~\citep{sun2024enhancing}, Reprompting~\citep{xu2023reprompting}, and bootstrapped demonstrations in DSPy~\citep{khattab2023dspy} add worked examples (F2); AWM~\citep{wang2024agent} and Voyager write reasoning procedures or workflows (F3); whereas TextGrad, GEPA, and OPRO rewrite the skill in full (F4). Figure~\ref{fig:design_space}(a) organizes these choices into three \emph{search strategies} (direct revision, iterative refinement, and parallel sampling) and four \emph{revision forms} (rules, examples, procedures, and rewrites). We call a search strategy together with the revision forms applied under it a \emph{revision operator}, and the methods above each commit to a single one. Some prompt optimizers choose among editing operators adaptively~\citep{ashizawa2025bandit,cui2025see}, though within one fixed search procedure.

\textbf{No single revision operator is a global winner.} We benchmark five of these methods under the same training budget. Figure~\ref{fig:design_space}(b) shows that the best baseline shifts with the benchmark and the target LLM: none of the existing methods ranks first in more than four of the eight benchmark-model combinations.
The same instability appears within \method when its search strategy is fixed. LiveMath accuracy drops to 39.34 with direct revision (I1) alone and 29.38 with iterative refinement (I2) alone, against 65.88 when the strategy is chosen adaptively (Table~\ref{tab:ablation_generation}).
\textbf{The best revision operator also changes as optimization proceeds.} Each accept or reject reshapes what the skill still needs. An accepted candidate skill removes the failures it was written for, so the operator that just succeeded may not suit the failures that remain. A rejected one leaves the skill unchanged but signals that this kind of change is not what the remaining failures need. Trace2Skill observes the same under greedy patching, where a patch that fixes some tasks breaks others and a patch for an already covered failure only restates existing guidance~\citep{ni2026trace2skill}.

We introduce \textbf{\method} (\textbf{stra}tegy-guided skill \textbf{tun}ing), which folds the choice of revision operator into the optimization (Figure~\ref{fig:motivation}). Each round, the frozen optimizer LLM reads the optimization state, which is the current execution feedback together with the recorded outcomes of earlier strategies and forms, then selects a search strategy and the revision forms used within it, and generates candidate skills (Figure~\ref{fig:motivation}(a)). Each strategy and form on this menu has a published precedent, shown in Figure~\ref{fig:design_space}(a)~\citep{yuksekgonul2024textgrad,agrawal2026gepa,yang2024large,zhao2024expel}; our contribution is the menu itself and the per-round selection from it. Candidate skills from all three search strategies and four revision forms pass through one candidate evaluation, and every accept or reject is written back into the optimization state, so the next choice of strategy and form builds on what earlier choices achieved (Figure~\ref{fig:motivation}(b)). Promising candidate skills are saved along the way, and final skill selection chooses among them and the current skill (Figure~\ref{fig:motivation}(c)).

Across four challenging benchmarks and two model settings, \method ranks first in most (Table~\ref{tab:main_results}), exceeding the strongest baseline by at least 3 points in three settings and by 46\% on LiveMath with Haiku. Ablations attribute the gain to adaptive selection, since random and fixed strategy choices lower scores on all four benchmarks.

We make three contributions. \textbf{(i) A design space of skill learning.} We organize existing methods by search strategy and revision form (Figure~\ref{fig:design_space}) and compare five of them under one training budget, finding that no single revision operator is best across benchmarks and target LLMs. \textbf{(ii) \method.} The optimizer LLM chooses the search strategy and revision form every round from the optimization state, and one candidate evaluation assesses all candidate skills under common criteria and records their outcomes for later choices. \textbf{(iii) Evidence that adaptive selection drives the gain.} \method ranks first in most benchmark-model settings and outperforms random, scheduled, bandit, and fixed strategy choices. Its choice of operator differs across benchmarks and keeps shifting within a run (Section~\ref{sec:experiments:discussion}).

%% file: sections/problem_statement.tex
\section{Problem Statement}
\label{sec:problem}
\label{sec:method:overview}
\label{sec:setup}

\paragraph{Skill learning.}
A large language model performs a task under a \emph{skill}, reusable text in its context. \emph{Skill learning} improves this text from the model's own executions while its parameters stay frozen. In each \emph{round}, the model executes a batch of training tasks under the current skill, a second language model (or the same) revises the skill from the scored executions, and the scores of the revised skill on further tasks decide whether it replaces the current one. 

\paragraph{Formulation.}
Let the \emph{target LLM} $F_\theta$ be the frozen model that performs the task, and the \emph{optimizer LLM} $G_\phi$ the frozen model that rewrites and optimizes the skill. Given an input $x\in\train$ from a training set and a skill $s$, the target LLM produces a \emph{trajectory} $\tau\sim F_\theta(\cdot\mid x,s)$, which is a response or a multi-step interaction. An \textit{evaluator} $r$ assigns it a score $r(x,\tau)\in[0,1]$. The quality of a skill $s$ is its expected score over the task distribution,
\begin{equation}
J(s)=\mathbb{E}_{x}\big[r(x,\tau)\big].
\label{eq:objective}
\end{equation}
Given $\train$, an \emph{initial skill} $s_0$, and a \emph{training budget} $B$ (defined below), skill learning seeks a \emph{final skill} $\hat{s}$ with $J$ as high as possible, and $J(\hat{s})$ is estimated on a held-out set $\test$. Following the common practice of mini-batch training, here we split $\train$ into $\numbatch$ batches. 
At each round $t$, $F_\theta$ executes the \emph{current skill} $s_t$ on one batch $\batch_t$, and these executions with their scores form the \emph{execution feedback} $E_t$. Based on $s_t$ and $E_t$, $G_\phi$ writes \emph{candidate skills} $c$. A \emph{candidate evaluation} then executes each candidate $c$ on training samples outside of $\batch_t$, compares its scores with those of $s_t$ on the same samples, and decides whether the candidate replaces $s_t$ as $s_{t+1}$. 
The main cost of skill learning is the executions of $F_\theta$, and search strategies differ in how many executions a round uses, so stopping after a fixed number of rounds or epochs would give methods unequal compute. We therefore define the budget $B$ as the total number of executions of $F_\theta$ on training samples, counting those for feedback and those for evaluating candidate skills. Test-set evaluation is not counted.

\paragraph{Our task.}
We study how candidate skills should be generated from $(s_t,E_t)$ at each round. We call the procedure that produces them the \emph{revision operator}. It combines a \emph{search strategy}, which determines how candidates are generated, with \emph{revision forms}, which determine what kind of text is written (three strategies I1--I3 and four forms F1--F4, Section~\ref{sec:method:strategies}). Existing methods fix one operator for the whole run, chosen before any feedback is seen. We instead choose the operator adaptively at every round. The operator $o_t$ at round $t$ is the search strategy used in that round together with the revision forms applied under it. It is chosen from the \emph{optimization state} $\Omega_t=(s_t,E_t,H_t,p_t)$. Besides the current skill and its execution feedback, $\Omega_t$ holds the \emph{evaluation history} $H_t$, which records the outcomes of earlier candidates grouped by strategy and by form, and $p_t$, which tracks a skill refinement in progress (see I2 in Section~\ref{sec:method:strategies}). The task is to choose $o_t$ from $\Omega_t$ at every round, with $\phi$ fixed and no selection policy trained, so that $\hat{s}$ attains the highest $J$ within $B$.

%% file: sections/method.tex
\section{Method}
\label{sec:method}

\begin{figure}[t]
    \centering
    \includegraphics[width=\linewidth]{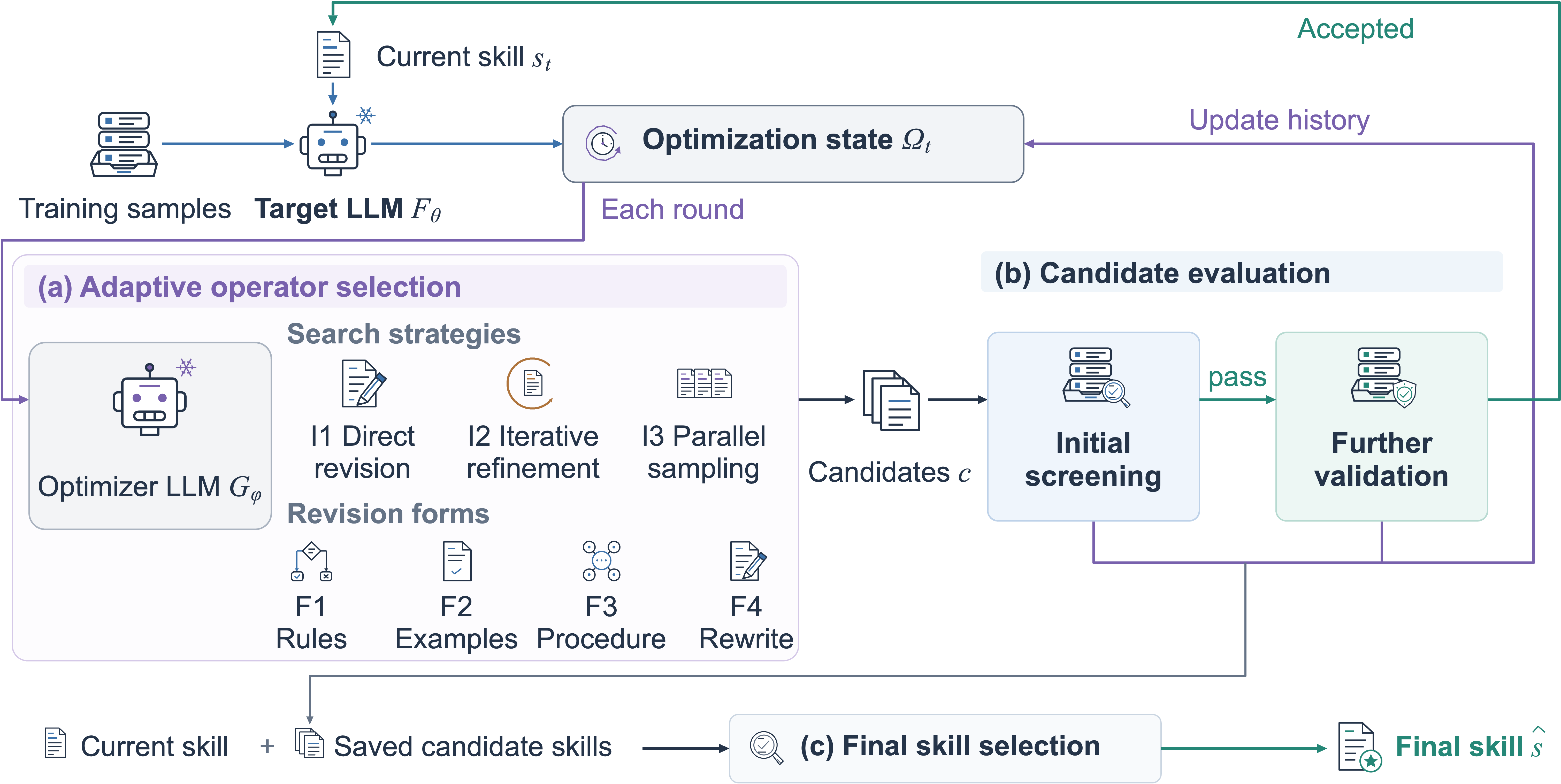}
    \caption{\textbf{Adaptive skill optimization with \method.} (a) From optimization state $\Omega_t$, optimizer $G_\phi$ adaptively selects a revision operator $o_t$ (a search strategy and the revision forms applied under it) and generates candidate skills, using execution feedback from target $F_\theta$ and task scores. (b) Candidate skills pass initial screening and then further validation; both stages record every outcome, including rejections, in the evaluation history $H_t$, and only candidates passing both replace the current skill. (c) After optimization, final skill selection compares the current skill with saved candidate skills to obtain $\hat{s}$.}
    \label{fig:motivation}
\end{figure}

Figure~\ref{fig:motivation} gives the overview. At round $t$, $G_\phi$ chooses the revision operator from $\Omega_t$ and generates candidate skills (Section~\ref{sec:method:selection}). A shared \emph{candidate evaluation} compares them with those of $s_t$ on training samples and records every outcome in $H_{t+1}$. After optimization, the \emph{final skill selection} chooses $\hat{s}$ among the current skill and the \emph{saved candidate skills} (Section~\ref{sec:method:evaluation}).

\subsection{Search Strategies and Revision Forms}
\label{sec:method:strategies}

Each strategy applies a form, so we describe the forms first. We consider four revision forms from existing work (Figure~\ref{fig:motivation}(a)): 
(1) \emph{Conditional rules (F1)} give instructions for particular situations~\citep{fu2024autoguide}; (2) \emph{Worked examples (F2)} pair a sample with its correct answer or a worked solution (i.e., few-shot examples~\citep{brown2020language}); (3) A \emph{reasoning procedure (F3)} specifies reasoning steps or a workflow of task operations~\citep{zhou2024self}; (4) A \emph{full rewrite (F4)} revises the entire skill to incorporate new guidance and reorganize existing instructions. Details in Appendix~\ref{app:strategies_forms}.

We define three alternative search strategies. (1) \emph{Direct Revision (I1)} uses $G_\phi$ to propose $c$ from the current skill $s_t$, choosing a revision form based on execution feedback $E_t$. (2) \emph{Iterative Refinement (I2)} maintains an \emph{intermediate skill} $\tilde{s}_t$, a separate working version initialized from $s_t$ when refinement starts. At each step, $G_\phi$ proposes $c$ by applying a selected revision form to $\tilde{s}_t$. An evaluated candidate skill may become the next intermediate skill even if it does not qualify to replace $s_t$. Refinement can therefore progress across rounds while the current skill remains unchanged. (3) \emph{Parallel Sampling (I3)} uses $G_\phi$ to generate $K$ candidate skills $c_1,\ldots,c_K$ from the same $s_t$ under one selected revision form. It ranks these alternatives on a small training subset before submitting selected candidate skills for evaluation.

\subsection{Adaptive Selection of Revision Operators}
\label{sec:method:selection}

Given these strategies and forms, the question at each round is which revision operator to apply. \method lets $G_\phi$ choose the operator $o_t$ from the optimization state $\Omega_t$. The execution feedback $E_t$ shows what the current skill still gets wrong, and the evaluation history $H_t$ shows how earlier revisions with each strategy and form fared on similar failures. Both matter, because a strategy that failed on one kind of failure may suit another, and a strategy that succeeded may stop helping once the skill has changed.

\paragraph{Using the optimization state.}
The evaluation history $H_t$ contains a \emph{strategy history}, grouping candidate outcomes by I1--I3, and a \emph{form history}, grouping them by explicitly selected revision forms. Both record, for each candidate skill, the attempt, the acceptance decision, the score change, and the \emph{regressions}, losses on previously solved samples. They also record whether a candidate skill's gain on the small screening set held up on the larger validation set (the two stages of candidate evaluation, Section~\ref{sec:method:evaluation}), which tells $G_\phi$ how far the gains of earlier revisions generalized.

For I2, $p_t$ records whether refinement has started and how many steps have been completed. This lets $G_\phi$ consider continuing work on $\tilde{s}_t$, even when no candidate skill has yet replaced $s_t$.

\paragraph{Operator selection and candidate generation.}
$G_\phi$ is prompted to choose an available strategy suited to $E_t$ and explain how the results in $H_t$ support its choice. It also selects revision forms and generates candidate skills under that strategy. Forms can change across successive refinement steps in I2. The first round uses I1, since no evaluation history exists yet. Appendix~\ref{app:adaptive_selection} gives the abbreviated selection prompt.

\subsection{Candidate Evaluation and State Updates}
\label{sec:method:evaluation}

Candidate evaluation decides whether a candidate $c$ replaces $s_t$ and, through its outcomes, supplies the evidence in $H_{t+1}$ for later operator choices. Two failure modes shape its design. A positive mean gain can hide regressions on samples the current skill already solves, and a gain measured on a few samples can vanish on more. Candidate evaluation therefore checks regressions explicitly and confirms gains first on a few samples and then on more.

\paragraph{Screening and validation.}
Each candidate skill is \emph{submitted} to candidate evaluation as soon as it is generated, rather than at the end of the round. All submitted candidate skills are compared with $s_t$ under the same procedure (Figure~\ref{fig:motivation}(b)). $F_\theta$ executes both skills on the same training subset $Q\subseteq\train$, and two measures decide the outcome. The first is the mean score gain of $c$ over $s_t$, written $\widehat{\Delta}_Q(c,s_t)$. The second is the number of regressions, samples that $s_t$ solves but $c$ fails or scores substantially lower on. A candidate with more regressions than improvements is rejected.

We employ two evaluation stages, both of which apply this comparison and differ only in the sample set and the minimum gain: \emph{1) Initial screening} uses a small set $Q_{\mathrm{s}}\subseteq\train$ that includes previously solved samples and excludes samples from the current batch $\batch_t$ that are used to generate $c$, and requires $\widehat{\Delta}_{Q_{\mathrm{s}}}(c,s_t)\geq\epsilon_{\mathrm{s}}$; 2) \emph{Further validation} is only applied to passing candidate skills on a larger \emph{validation set} $Q_{\mathrm{v}}\subseteq\train$, a random sample of training samples disjoint from $Q_{\mathrm{s}}$ and from $\batch_t$; no separate validation split is used. Validation requires $\widehat{\Delta}_{Q_{\mathrm{v}}}(c,s_t)\geq\epsilon_{\mathrm{v}}$ against the same $s_t$ and tests whether the screening gain persists. Exact criteria are given in Appendix~\ref{app:evaluation_rules}.

\paragraph{Updating the optimization state.}
Only a candidate skill passing both stages replaces $s_t$ as the current skill. Every outcome, including rejection, updates the evaluation history $H_{t+1}$, that is, the strategy and form histories. These histories record which strategies and forms were tried, their observed gains and regressions, and whether screening gains persisted. Together with the skill used for the next batch, its execution feedback, and refinement progress, $H_{t+1}$ forms $\Omega_{t+1}$. Rejected candidate skills therefore still supply evidence for later operator choices, even though they leave the current skill unchanged.

\paragraph{Final skill selection.}
Saved candidate skills include accepted versions and screened candidates whose gain was positive but below the acceptance threshold. After optimization, final skill selection compares saved candidate skills with the current skill on a common set of training samples (Figure~\ref{fig:motivation}(c)) and keeps a candidate only if it passes the gain and regression checks. The resulting skill is the final skill $\hat{s}$; Appendix~\ref{app:evaluation_rules} gives details.

%% file: sections/experiments.tex
\section{Experiments}
\label{sec:experiments}
Our experiments answer four research questions (RQ). \textbf{RQ1} Does selecting the revision operator adaptively at each round outperform methods that commit to one? \textbf{RQ2} What does the per-round choice of strategy and form contribute, and what does it cost? \textbf{RQ3} Which revision operators does \method choose, and how do the skills evolve during optimization? \textbf{RQ4} Do the learned skills generalize to a stronger target LLM?

\subsection{Experimental Setup}
\label{sec:experiments:setup}

\paragraph{Tasks and model settings.}
We evaluate \method and five recent baselines on four real-world datasets: 1) DocVQA~\citep{mathew2021docvqa} for document understanding, 2) LiveMathematicianBench (LiveMath)~\citep{he2026livemathematicianbench} for mathematical reasoning, 3) Mind2Web~\citep{deng2023mind2web} for offline element selection conditioned on annotated action histories, and 4) SpreadsheetBench~\citep{ma2024spreadsheetbench} for spreadsheet manipulation. Following previous works, we report average normalized Levenshtein similarity (ANLS) for DocVQA~\citep{biten2019scene,mathew2021docvqa}, answer accuracy for LiveMath~\citep{he2026livemathematicianbench}, element accuracy averaged over tasks for Mind2Web~\citep{deng2023mind2web,wang2024agent}, and the fraction of tasks passing all workbook checks for SpreadsheetBench~\citep{ma2024spreadsheetbench}. The main model setting uses Claude Haiku~4.5 as the target LLM and Claude Sonnet~4.6 as the optimizer LLM. We also evaluate another setting with a Claude Opus~4.8 target LLM and a Claude Opus~5 optimizer. All model parameters remain frozen.

\paragraph{Baselines and training budget.}
We compare \method with five self-evolving skill baselines: TextGrad~\citep{yuksekgonul2024textgrad}, GEPA~\citep{agrawal2026gepa}, SkillOpt~\citep{yang2026skillopt}, Trace2Skill~\citep{ni2026trace2skill}, and Iter-CoT~\citep{sun2024enhancing}. Within each dataset and model setting, all methods share the same target and optimizer LLMs, initial skill, task interface, and test samples. Prior work shows that differences in the compute spent during optimization can affect comparisons of skill and memory modules~\citep{hajimiri2026online}. Each run therefore allows $6|\train|$ target-LLM executions as the total training budget $B$ (Section~\ref{sec:problem}), including final skill selection and excluding optimizer calls and test evaluation. 

We also present ablations of \method in Section~\ref{sec:experiments:main}. We change one element of \method per run under the same training budget. \emph{I1 only}, \emph{I2 only}, and \emph{I3 only} fix the search strategy, \emph{Rand.} samples a strategy at random each round, and \emph{Rot.} cycles through I1, I2, and I3 in consecutive rounds. \emph{Once} keeps the strategy that the optimizer LLM chooses at its first free decision, \emph{Replay} follows round by round the strategy sequence recorded in the \method run of Table~\ref{tab:main_results} in a new run with a different order of training batches, and \emph{No hist.} withholds the strategy and form histories $H_t$ from the strategy selection. \emph{Bandit} replaces the optimizer LLM in the strategy choice by a multi-armed bandit rule over the outcomes recorded in $H_t$, either $\epsilon$-greedy~\citep{sutton1998reinforcement}, UCB1~\citep{auer2002finite}, or Thompson sampling~\citep{william1933likelihood}. In the above ablation settings, the optimizer LLM still chooses the revision form adaptively, while \emph{F4 only} keeps adaptive strategy selection but restricts every candidate skill to a full rewrite. We also conduct ablation studies on the candidate skill evaluation method: \emph{No valid.} accepts candidate skills after initial screening alone, and \emph{No final sel.} takes the current skill at the end of optimization as the final skill.

\paragraph{Implementation details.}
\label{sec:experiments:implementation}
Training samples are split into $\numbatch=16$ batches with the same mix of task types, and each round uses one batch. To pass initial screening, a candidate skill must raise the mean score over the current skill by at least $\epsilon_{\mathrm{s}}$ (at least 0.001 on DocVQA and 0.0025 elsewhere) and pass the regression checks of Appendix~\ref{app:evaluation_rules}. Further validation applies its own gain and regression checks on a larger sample set, with $\epsilon_{\mathrm{v}}=0.01$. The calls to $F_\theta$ and $G_\phi$ use temperature zero to maintain reproducibility. Further details of datasets, baselines, and ablations are in Appendix~\ref{app:experimental_details}. We report variability across four runs with different random seeds of \method in Appendix~\ref{app:run_variability}.

\subsection{Main Results}
\label{sec:experiments:main}

\begin{table}[!t]
\centering
\small
\caption{Test scores (0--100) with Haiku and Opus as target LLMs under a training budget of $6|\train|$ target-LLM executions. Bold marks the highest score in each model setting.}
\label{tab:main_results}
\begin{tabular*}{\linewidth}{@{\extracolsep{\fill}}lrrrr@{}}
\toprule
Method & DocVQA & LiveMath & Mind2Web & SpreadsheetBench \\
\midrule
\multicolumn{5}{l}{\textbf{Haiku setting}\quad Target: Haiku 4.5 \quad Optimizer: Sonnet 4.6} \\
\addlinespace[2pt]
Initial skill & 47.88 & 29.38 & 46.09 & 40.00 \\
GEPA        & 91.26 & 43.60 & 46.18 & 52.50 \\
TextGrad    & 90.69 & 31.75 & 46.16 & 45.00 \\
SkillOpt    & 91.87 & 45.02 & 40.55 & 50.71 \\
Trace2Skill & 90.62 & 33.18 & 46.18 & 46.79 \\
Iter-CoT    & 73.42 & 36.97 & \textbf{47.99} & 48.57 \\
\method     & \textbf{92.12} & \textbf{65.88} & 47.49 & \textbf{56.43} \\
\midrule
\multicolumn{5}{l}{\textbf{Opus setting}\quad Target: Opus 4.8 \quad Optimizer: Opus 5} \\
\addlinespace[2pt]
Initial skill & 94.04 & 42.18 & 41.80 & 39.29 \\
GEPA        & 95.26 & 78.20 & 50.50 & \textbf{72.86} \\
TextGrad    & 94.37 & 52.61 & 50.32 & 71.43 \\
SkillOpt    & 95.25 & 74.41 & 50.26 & 70.71 \\
Trace2Skill & 95.56 & 53.08 & 50.20 & 71.79 \\
Iter-CoT    & 95.25 & 48.34 & 49.92 & 65.71 \\
\method     & \textbf{96.95} & \textbf{81.52} & \textbf{51.63} & \textbf{72.86} \\
\bottomrule
\end{tabular*}
\end{table}

\begin{table}[!t]
\centering
\small
\caption{Test scores of ablations of candidate generation with Haiku/Sonnet (defined in Section~\ref{sec:experiments:setup}).}
\label{tab:ablation_generation}
\begin{tabular*}{\linewidth}{@{\extracolsep{\fill}}lrrrr@{}}
\toprule
Configuration & DocVQA & LiveMath & Mind2Web & SpreadsheetBench \\
\midrule
\method & \textbf{92.12} & \textbf{65.88} & \textbf{47.49} & \textbf{56.43} \\
\addlinespace[2pt]
\multicolumn{5}{l}{\textit{Fixed search strategy}} \\
\quad \emph{I1 only} & 91.92 & 39.34 & 46.09 & 54.64 \\
\quad \emph{I2 only} & 91.56 & 29.38 & 47.34 & 52.14 \\
\quad \emph{I3 only} & 90.73 & 63.03 & 45.99 & 55.00 \\
\addlinespace[2pt]
\multicolumn{5}{l}{\textit{Strategy chosen without the per-round decision}} \\
\quad \emph{Rand.} & 91.10 & 63.98 & 46.58 & 52.50 \\
\quad \emph{Rot.} & 91.74 & 65.40 & 46.06 & 52.14 \\
\quad \emph{Once} & 91.53 & 63.03 & 46.14 & 55.36 \\
\quad \emph{Replay} & 92.00 & 51.66 & 46.18 & 53.21 \\
\addlinespace[2pt]
\multicolumn{5}{l}{\textit{Strategy chosen by a bandit rule over the histories}} \\
\quad \emph{Bandit}, $\epsilon$-greedy & 91.94 & 55.45 & 46.09 & 49.64 \\
\quad \emph{Bandit}, UCB1 & 90.92 & 38.39 & 46.02 & 53.21 \\
\quad \emph{Bandit}, Thompson & 91.41 & 45.02 & 46.07 & 53.57 \\
\addlinespace[2pt]
\multicolumn{5}{l}{\textit{Strategy chosen without the histories}} \\
\quad \emph{No hist.} & 91.74 & 34.12 & 46.09 & 52.14 \\
\addlinespace[2pt]
\multicolumn{5}{l}{\textit{Fixed revision form}} \\
\quad \emph{F4 only} & 90.43 & 42.18 & 46.07 & 51.79 \\
\bottomrule
\end{tabular*}
\end{table}

\begin{table}[!t]
\centering
\small
\caption{Test scores of ablations of candidate evaluation with Haiku/Sonnet (defined in Section~\ref{sec:experiments:setup}).}
\label{tab:ablation_evaluation}
\begin{tabular*}{\linewidth}{@{\extracolsep{\fill}}lrrrr@{}}
\toprule
Configuration & DocVQA & LiveMath & Mind2Web & SpreadsheetBench \\
\midrule
\method & \textbf{92.12} & \textbf{65.88} & \textbf{47.49} & \textbf{56.43} \\
\emph{No valid.} & 91.51 & 40.28 & 46.09 & 47.86 \\
\emph{No final sel.} & \textbf{92.12} & \textbf{65.88} & \textbf{47.49} & 51.43 \\
\bottomrule
\end{tabular*}
\end{table}

\begin{table}[!t]
\begin{minipage}[t]{0.56\linewidth}
\centering
\footnotesize
\setlength{\tabcolsep}{4pt}
\caption{Test scores with Haiku/Sonnet. The indented rows keep the candidate generation of GEPA or SkillOpt but use the candidate evaluation of \method.}
\label{tab:hybrid_generation}
\begin{tabular}{lrr}
\toprule
Configuration & LiveMath & \multicolumn{1}{r}{\shortstack[r]{Spreadsheet\\Bench}} \\
\midrule
GEPA & 43.60 & 52.50 \\
\quad with \method evaluation & 37.91 & 39.64 \\
SkillOpt & 45.02 & 50.71 \\
\quad with \method evaluation & 52.13 & 53.93 \\
\method & \textbf{65.88} & \textbf{56.43} \\
\bottomrule
\end{tabular}
\end{minipage}\hfill
\begin{minipage}[t]{0.41\linewidth}
\centering
\vspace{0pt}
\includegraphics[width=\linewidth]{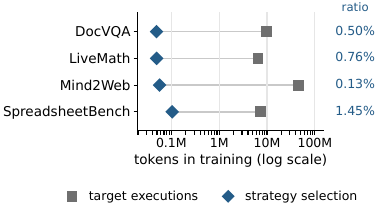}
\captionof{figure}{Training tokens of \method with Haiku/Sonnet on target-LLM executions (squares) and strategy selection (diamonds), and the latter as a percentage of the former.}
\label{fig:token_cost}
\end{minipage}
\end{table}

\paragraph{Comparison with baselines (RQ1).}
\method is the best-performing method in Table~\ref{tab:main_results}. It achieves the highest score in six of the eight dataset--model settings, has the same highest score as GEPA on SpreadsheetBench with Opus, and is the second best on Mind2Web with Haiku. With Haiku/Sonnet, its largest gain is on LiveMath, where it scores 65.88 compared with 45.02 for SkillOpt. It also improves on the strongest baselines by 3.93 points on SpreadsheetBench and 0.25 points on DocVQA, and on Mind2Web it trails Iter-CoT by 0.50 points, with GEPA and Trace2Skill at 46.18. With Opus, \method exceeds the strongest baselines by 1.39, 3.32, and 1.13 points on DocVQA, LiveMath, and Mind2Web, respectively, and matches GEPA at 72.86 on SpreadsheetBench. The baselines are less consistent. The strongest baseline is GEPA, which achieves the best scores across all baselines in four settings, SkillOpt in two, and Iter-CoT and Trace2Skill in one each. Iter-CoT, the strongest on Mind2Web with Haiku, is the weakest on DocVQA with Haiku and on LiveMath, Mind2Web, and SpreadsheetBench with Opus. On DocVQA, the initial skill already scores 94.04 with Opus against 47.88 with Haiku, leaving less room for improvement. Taken together, each baseline commits to a single revision operator (Figure~\ref{fig:design_space}), and none of them performs consistently well across all tasks, whereas \method, which selects the revision operator adaptively, is the only method that ranks first or within 0.5 points of first in all eight settings.

\begin{figure}[!t]
\centering
\includegraphics[width=0.9\linewidth]{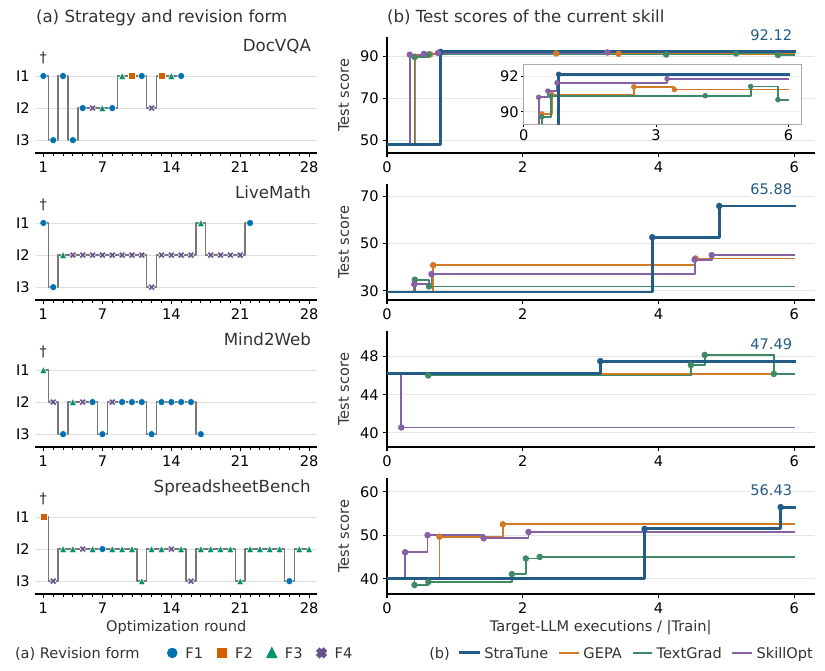}
\caption{The \method runs of Table~\ref{tab:main_results} with Haiku/Sonnet. \textbf{(a)} Strategy and revision form (marker) chosen in each round and $\dagger$ marks initialization. \textbf{(b)} Test score of the current skill against the number of $F_\theta$ executions divided by $|\train|$; filled circles mark changes of the current skill.}
\label{fig:adaptive_optimization}
\end{figure}

\paragraph{Ablation of adaptive candidate generation (RQ2).}
Table~\ref{tab:ablation_generation} compares \method with the ablations of candidate generation defined in Section~\ref{sec:experiments:setup}, all with the same candidate evaluation and training budget. \textbf{The best fixed search strategy depends on the task.} I1 alone is strongest on DocVQA, I3 on LiveMath and SpreadsheetBench, and I2 on Mind2Web. \method exceeds each fixed strategy on all four datasets. \textbf{Removing the per-round decision lowers the score on every dataset.} \emph{Rand.} is below \method by 1.02, 1.90, 0.91, and 3.93 points. \emph{Rot.} stays within 0.5 points of \method on DocVQA and LiveMath and trails by 1.43 and 4.29 points on Mind2Web and SpreadsheetBench. \emph{Once} trails by 0.59, 2.85, 1.35, and 1.07 points; on LiveMath, its chosen strategy was I3, so that run equals the I3-only score. \emph{Replay} is below \method on every dataset, by 0.12, 14.22, 1.31, and 3.22 points. Its run draws different training batches and candidate skills, to which the copied sequence does not respond. \textbf{A bandit rule over the outcomes does not replace the optimizer LLM.} Choosing the strategy by $\epsilon$-greedy, UCB1, or Thompson sampling from the same recorded outcomes scores below \method on all four datasets, by at most 1.5 points on DocVQA and Mind2Web and by 2.9 to 27.5 points on LiveMath and SpreadsheetBench. With at most two accepted candidate skills per run, their reward is almost always zero, so they either keep exploring or lock onto the first rewarded strategy, whereas the optimizer LLM also reads what the current skill gets wrong (Appendix~\ref{app:bandit_selectors}). \textbf{The histories inform the decision.} Withholding them from the strategy selection (\emph{No hist.}) lowers the score by 0.38, 31.76, 1.40, and 4.29 points, most on LiveMath, where the run falls to 34.12. \textbf{The choice of revision form contributes on its own.} \emph{F4 only} lowers the score on every dataset by 1.69, 23.70, 1.42, and 4.64 points, most on LiveMath, so full rewrites alone miss improvements that rules, examples, and procedures express, and the choice of form should also stay with the optimizer LLM. Appendix~\ref{app:fixed_operators} fixes both at once.

Although some ablations come close on individual datasets, \method ranks first on all four benchmarks in Table~\ref{tab:ablation_generation}. The strongest ablation differs across datasets, suggesting that adaptive selection helps maintain consistently strong performance across tasks.

\paragraph{Ablation of candidate evaluation and final skill selection (RQ2).}
Table~\ref{tab:ablation_evaluation} shows that both evaluation steps matter. Accepting candidate skills after initial screening alone (\emph{No valid.}) lowers the score on all four datasets, most on LiveMath, from 65.88 to 40.28, and on SpreadsheetBench, from 56.43 to 47.86. Final skill selection confirms the current skill on three datasets and on SpreadsheetBench selects a candidate skill proposed in round 2 as the final skill, raising the score from 51.43 to 56.43 (\emph{No final sel.}).

\paragraph{Baseline candidate generation with the candidate evaluation of \method (RQ2).}
Table~\ref{tab:hybrid_generation} isolates the contribution of candidate evaluation. The two indented rows keep the candidate evaluation and final skill selection of \method and the same training budget, and differ only in who generates the candidate skills, GEPA or SkillOpt (Appendix~\ref{app:baseline_settings}). In both cases the same evaluation yields a weaker skill than \method, so the gain cannot be attributed to evaluation alone. GEPA's reflective rewrites mostly fail the evaluation, so the current skill is rarely replaced and the scores fall below GEPA itself. This measures only GEPA's proposal step under our candidate evaluation; Table~\ref{tab:main_results} compares GEPA's own search. SkillOpt's small edits pass more often, improving its LiveMath score from 45.02 to 52.13. Its scores on both datasets remain below \method and, on SpreadsheetBench, below \emph{I1 only}. Candidate evaluation thus decides which candidate skills are kept, while the gain is set by how they are generated.

\paragraph{Token overhead of strategy selection (RQ2).}
Existing self-evolving skill methods spend their training cost on generating candidate skills with the optimizer LLM and evaluating them with the target LLM~\citep{gao2025survey}. \method adds one optimizer call per round for strategy selection and no target-LLM executions, and we measure this added cost in tokens on complete Haiku/Sonnet runs (Figure~\ref{fig:token_cost}). The strategy-selection calls use 0.50, 0.76, 0.13, and 1.45 percent of the tokens spent on target-LLM executions on DocVQA, LiveMath, Mind2Web, and SpreadsheetBench. Strategy selection therefore adds a small token overhead relative to target-LLM executions. Appendix~\ref{app:token_accounting} reports all training tokens used by the target LLM and optimizer LLM in these runs.

\subsection{Discussion}
\label{sec:experiments:discussion}

\paragraph{Strategy and form choices (RQ3).}
Figure~\ref{fig:adaptive_optimization}(a) shows the strategy and revision form \method chose in each round; the choice settles into a task-specific pattern yet keeps shifting within it. On LiveMath, 16 of the 22 rounds refine the intermediate skill through full rewrites (I2 with F4), whereas on SpreadsheetBench the same strategy edits reasoning procedures (I2 with F3) in 19 of 28 rounds. On DocVQA, direct revision (I1) is chosen in 8 of 15 rounds, proposing conditional rules (F1) in 4 of them and worked examples or reasoning procedures (F2 or F3) in the others, so the form varies even under one strategy. On LiveMath, Mind2Web, and SpreadsheetBench, parallel sampling (I3) appears in short bursts between longer sequences of I2 rounds, and the form it samples follows the task, conditional rules (F1) on Mind2Web and mostly full rewrites or reasoning procedures (F4 or F3) on SpreadsheetBench. \textbf{No two benchmarks follow the same sequence}.

\paragraph{Optimization trajectories (RQ3).}
Figure~\ref{fig:adaptive_optimization}(b) plots the test score of the current skill after $k|\train|$ target-LLM executions. The baselines obtain most of their gain within the first $|\train|$ executions, and their later updates are small and sometimes harmful, as when SkillOpt's Mind2Web score drops to 40.55 at $0.21|\train|$ and never recovers. \method accepts one or two updates per benchmark and each raises the test score, although they arrive later, after $3|\train|$ or more on LiveMath, Mind2Web, and SpreadsheetBench. They arrive later because a candidate skill must pass both initial screening and further validation (Section~\ref{sec:method:evaluation}).

\begin{figure}[!t]
\centering
\includegraphics[width=\linewidth]{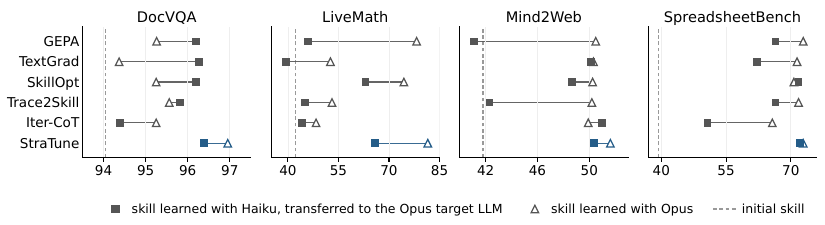}
\caption{Test scores with Opus~4.8 as the target LLM of skills learned with Haiku (squares) and with Opus itself (triangles). The dashed line is the initial skill, and \method is highlighted by blue.}
\label{fig:opus_transfer}
\end{figure}

\paragraph{Transfer to a stronger target LLM (RQ4).}
Final skills learned with Haiku also improve a stronger target LLM without further optimization. Figure~\ref{fig:opus_transfer} evaluates the Haiku final skills of every method in Table~\ref{tab:main_results} with Opus~4.8 as the target. The \method skills raise the score of the initial skill on every dataset and recover 81, 60, 87, and 98 percent of the gain of the \method skills learned with Opus itself. The transfer is weakest on LiveMath, where the Haiku skill mainly teaches one recurring type of question and gains nothing on the rest, whereas the skill learned with Opus also improves them. On DocVQA, the Haiku skills of GEPA, TextGrad, SkillOpt, and \method score above every baseline skill learned with Opus itself. All four contain one rule, to answer with the value alone, for example \emph{Dallas} rather than \emph{Dallas, Texas}, because such format errors cause most failures of Haiku, whose initial skill scores 47.88. Opus starts from 94.04 and meets these errors too rarely in a training batch to learn the rule, yet they still make up two thirds of its remaining failures, so the rule learned with Haiku removes them. The changes these skills make therefore concern the task rather than the target LLM. \method can thus learn skills with an inexpensive target LLM and apply them to a stronger, more expensive one, reducing the cost of optimization.
\FloatBarrier

%% file: sections/conclusion.tex
\section{Conclusion}
\label{sec:conclusion}
We presented \method, which makes the choice of revision operator, a search strategy and the revision forms applied under it, part of skill optimization. The optimizer LLM selects the operator each round from the optimization state, and one candidate evaluation decides which candidates replace the current skill and records the outcome for later choices. \method ranks first in most of the eight benchmark-model settings, its ablations attribute the gain to choosing how to learn rather than to evaluation alone, and skills learned with Haiku also improve a stronger target LLM without further optimization.

%% file: sections/appendix.tex
\section{Search Strategies and Revision Forms}
\label{app:strategies_forms}

\subsection{Search Strategies}
\label{app:search_strategies}

Execution feedback can suggest several ways to improve a skill. The optimizer LLM $G_\phi$ may compare different changes, continue refining an intermediate skill, or generate several candidate skills within one revision form. We provide three search strategies to support these different needs. Each takes the current skill $s_t$ and the execution feedback $E_t$ from the optimization state $\Omega_t$ and submits candidate skills $c$ to candidate evaluation (Appendix~\ref{app:evaluation_rules}).

Direct Revision (I1) prompts $G_\phi$ with $s_t$ and $E_t$ to generate a candidate skill $c$, choosing the revision form that fits the observed failures, for example a conditional rule (F1) or a worked example (F2). The candidate skill proceeds to candidate evaluation.

Iterative Refinement (I2) retains a separate working version, the intermediate skill $\tilde{s}_t$, for continued revision across rounds. It is initialized from $s_t$ when refinement starts. Each refinement step applies one selected revision form to $\tilde{s}_t$: F1--F3 edit it locally, and F4 rewrites it. $G_\phi$ generates candidate skills from $\tilde{s}_t$, compares them on a common training subset, and submits higher-scoring candidates for candidate evaluation. An evaluated candidate skill may become the next intermediate skill $\tilde{s}_{t+1}$ even if it does not yet qualify to replace $s_t$. The progress variable $p_t$ in $\Omega_t$ records whether refinement has started and how many steps it has completed, and selecting I2 again resumes it, allowing several rounds of revision before a candidate skill is accepted.

Parallel Sampling (I3) generates $K$ candidate skills $c_1,\ldots,c_K$ from the same $s_t$ and $E_t$ under one selected revision form. Separate generation calls produce alternative wordings of the same kind of change. The $K$ candidates are first scored on a small shared set of training samples. The highest-scoring candidate skill proceeds to candidate evaluation, and so does the runner-up if its score on these samples is no more than 0.02 below the current skill's, since the small shared set is too noisy to discard a near tie. This helps when the value of a modification depends on how it is expressed.

\subsection{Revision Forms}
\label{app:revision_forms}

Table~\ref{tab:forms} defines the four revision forms. Their use within each search strategy is described in Section~\ref{sec:method:strategies}.

\begin{table}[htbp]
\centering
\small
\caption{Revision forms used to generate candidate skills.}
\label{tab:forms}
\begin{tabularx}{\linewidth}{@{}l p{0.25\linewidth}>{\raggedright\arraybackslash}X@{}}
\toprule
Form & Name & Content introduced or revised \\
\midrule
F1 & Conditional rules & Instructions for particular situations, such as converting quantities to a common unit before comparison. \\
\addlinespace
F2 & Worked examples & A sample paired with a correct answer or solution procedure that illustrates the intended behavior. \\
\addlinespace
F3 & Reasoning procedure & Steps for solving a task, including reasoning and answer checking or a workflow of task operations. Examples include comparing two derivations, or reading a table, filtering rows, and writing the result. \\
\addlinespace
F4 & Full rewrite & A rewrite of the entire skill to incorporate guidance, group related instructions, and resolve conflicts. \\
\bottomrule
\end{tabularx}
\end{table}
\FloatBarrier

\subsection{Adaptive Selection of Revision Operators}
\label{app:adaptive_selection}

Figure~\ref{fig:strategy_prompt} gives the abbreviated selection prompt used in Section~\ref{sec:method:selection}. It supplies execution feedback $E_t$, descriptions of I1--I3, the strategy and form histories in $H_t$, and refinement progress $p_t$.

\input{figures/strategy_prompt}

Under I1, $G_\phi$ selects a revision form while generating a candidate $c$ from $s_t$. Under I2 and I3, it selects one form together with the search strategy. For I2, the selected form is applied to $\tilde{s}_t$ in the current refinement step: F1--F3 edit it locally, while F4 rewrites it. Successive steps can therefore use different forms. For I3, $G_\phi$ generates $c_1,\ldots,c_K$ from $s_t$ using the selected form.

\section{Skill Evaluation Rules}
\label{app:evaluation_rules}

This section states the evaluation criteria summarized in Section~\ref{sec:method:evaluation}. Threshold schedules, early stopping, and maintenance of the pool of saved candidate skills follow the released implementation.

\paragraph{Paired comparison.}
Let $c$ be a candidate skill and $b$ the reference skill it is compared with; during candidate evaluation, $b=s_t$. The target LLM $F_\theta$ executes each sample $x$ in an evaluation set $Q$ under both skills. For an execution $\tau$ under skill $s$, the observed score is $\hat r(x;s)=r(x,\tau)$, with $r$ defined in Section~\ref{sec:setup}. The mean gain is
\begin{equation}
    \widehat{\Delta}_{Q}(c,b)
    =\frac{1}{|Q|}\sum_{x\in Q}
    \left[\hat r(x;c)-\hat r(x;b)\right].
    \label{eq:paired}
\end{equation}

\paragraph{Regressions.}
Let $\mathcal{S}_b,\mathcal{S}_c\subseteq Q$ be the sets of samples solved with $b$ and $c$ according to the dataset's success criterion, let $d_x=\hat r(x;c)-\hat r(x;b)$, and let $\delta=0.10$ mark a substantial change of a continuous score. We define
\begin{equation}
    S=|\mathcal{S}_b|,\;
    D=\left|(\mathcal{S}_b\setminus\mathcal{S}_c)
      \cup\{x\in\mathcal{S}_b:d_x\leq-\delta\}\right|,\;
    U=\left|(\mathcal{S}_c\setminus\mathcal{S}_b)
      \cup\{x\in Q:d_x\geq\delta\}\right|,
    \label{eq:outcome_counts}
\end{equation}
so $D$ counts newly failed samples or large decreases among previously solved samples, and $U$ counts newly solved samples or large increases; each sample is counted at most once in $D$ or $U$. Two loss proportions are checked: $D/S$, the share of previously solved samples on which the candidate regressed, and $D/(D+U)$, the share of losses among all changes. Screening sets hold only 36 to 48 samples, so these proportions are noisy, and we test them through a confidence bound rather than the raw ratio. We write $\mathrm{LB}(k,n)$ for the one-sided 90\% Wilson lower confidence bound~\citep{wilson1927probable} on the true proportion when $k$ of $n$ samples are losses: with $\hat p=k/n$ and $z=1.2816$, the 90\% quantile of the standard normal distribution,
\begin{equation}
    \mathrm{LB}(k,n)=\max\!\left(0,\;\frac{\hat p+\frac{z^2}{2n}-z\sqrt{\frac{\hat p(1-\hat p)}{n}+\frac{z^2}{4n^2}}}{1+\frac{z^2}{n}}\right),
    \qquad \mathrm{LB}(k,0)=0.
    \label{eq:wilson}
\end{equation}
$\mathrm{LB}(k,n)$ lies below $\hat p$ and approaches it as $n$ grows, so a few losses on a small set do not by themselves fail a candidate. The two regression checks are $\mathrm{LB}(D,S)\leq\tfrac12$ and $\mathrm{LB}(D,D+U)\leq\tfrac12$: a check fails only when even the lower bound says that losses are a majority, and a check with no eligible samples passes.

\paragraph{Initial screening.}
The screening set $Q_{\mathrm{s}}\subseteq\train$ combines samples previously solved with $b$ and a stratified random sample of the remaining eligible samples, excluding the training batch and any recorded examples used to generate $c$. The candidate skill passes if $\widehat{\Delta}_{Q_{\mathrm{s}}}(c,b)\geq\epsilon_{\mathrm{s}}$, the number of samples on which $c$ scores higher than $b$ is at least the number on which it scores lower, $\mathrm{LB}(D_{\mathrm{s}},S_{\mathrm{s}})\leq\tfrac12$, and $\mathrm{LB}(D_{\mathrm{s}},D_{\mathrm{s}}+U_{\mathrm{s}})\leq\tfrac12$, where the subscripts indicate the samples used. The minimum screening gain is $\epsilon_{\mathrm{s}}=\max(\epsilon_0,\tilde m/|Q_{\mathrm{s}}|)$, where $\epsilon_0$ is a configured floor (Appendix~\ref{app:experimental_details}) and $\tilde m$ is the median absolute nonzero score difference between $c$ and $b$ on $Q_{\mathrm{s}}$; for binary scores $\tilde m=1$, so the candidate must gain at least one net sample.

\paragraph{Further validation.}
For each candidate skill, the validation set $Q_{\mathrm{v}}$ is a random sample of $\mathrm{round}(0.3\,|\train|)$ training samples that excludes the candidate's screening samples and the batch used to generate it. Against the same reference skill $b$, the candidate skill must satisfy $\widehat{\Delta}_{Q_{\mathrm{v}}}(c,b)\geq\epsilon_{\mathrm{v}}$ and $\mathrm{LB}(D_{\mathrm{v}},D_{\mathrm{v}}+U_{\mathrm{v}})\leq\tfrac12$, where $\epsilon_{\mathrm{v}}$ is the minimum validation gain. Table~\ref{tab:stratune_settings} lists the dataset-specific set sizes.

\paragraph{Saved candidate skills.}
Accepted candidates are saved, including earlier accepted versions replaced by later updates. Candidates that pass the regression checks with a positive gain below $\epsilon_{\mathrm{s}}$ are also saved. 

\paragraph{Final skill selection.}
After optimization, the current skill and the saved candidate skills are executed on a common set of training samples, which is split into a selection subset (60\%) and a confirmation subset (40\%). The two subsets are disjoint so that a candidate chosen for its gain on the selection subset is confirmed on samples that did not influence the choice; choosing and confirming on the same samples favours candidates whose measured gain is partly noise. The current skill is the selected skill at the start. Each saved candidate skill in turn is compared with the selected skill and replaces it only if it passes on both subsets. On the selection subset, its mean gain over the selected skill must be at least $\max(0.01,\tilde m/n)$, where $n$ is the size of the subset and $\tilde m$ is defined as in initial screening, the number of samples on which it scores higher than the selected skill must be at least the number on which it scores lower, and $\mathrm{LB}(D,S)\leq\tfrac12$ and $\mathrm{LB}(D,D+U)\leq\tfrac12$ must hold. On the confirmation subset, the same conditions apply with a positive mean gain in place of the threshold. In this stage $D$ counts only samples solved by the selected skill and failed by the candidate, without the $\delta$ term of Equation~\ref{eq:outcome_counts}. The selected skill after the last comparison is the final skill $\hat{s}$.

\input{sections/experimental_details}
\FloatBarrier
\input{sections/appendix_results}

%% file: figures/strategy_prompt.tex
\begin{figure}[!t]
\centering
\begingroup
\definecolor{stratunePromptTitle}{HTML}{404040}
\definecolor{stratunePromptBackground}{HTML}{EEF2FA}
\setlength{\fboxsep}{0pt}
\setlength{\fboxrule}{0.5pt}
\fcolorbox{stratunePromptTitle}{stratunePromptBackground}{%
\begin{minipage}{\dimexpr\linewidth-2\fboxrule\relax}
\setlength{\fboxsep}{7pt}
\noindent\colorbox{stratunePromptTitle}{%
\parbox{\dimexpr\linewidth-2\fboxsep\relax}{%
\color{white}\normalsize\bfseries StraTune Revision Operator Selection Prompt}}%
\par\vspace{7pt}
\noindent\hspace*{7pt}%
\begin{minipage}{\dimexpr\linewidth-14pt\relax}
\normalfont\small\raggedright
You are optimizing a reusable skill for a target model. Choose a search strategy and, where applicable, a revision form for the next iteration using the execution feedback and history below.\par
\medskip
\textbf{Execution feedback}\par
\textit{[Summary of current failures, recent evaluation results, and the availability of supporting examples and successful trajectories.]}\par
\smallskip
\textbf{Strategy history}\par
\textit{[Previous attempts, acceptance decisions, and observed gains or regressions, grouped by search strategy.]}\par
\smallskip
\textbf{Form history}\par
\textit{[Previous attempts, acceptance decisions, and observed gains or regressions, grouped by explicitly selected revision forms.]}\par
\smallskip
\textbf{Refinement progress}\par
\textit{[Whether I2 has started and how many refinement steps have been completed.]}\par
\medskip
\textbf{Search strategies}\par
I1 Direct Revision (one form for the candidate); I2 Iterative Refinement (one form per step); I3 Parallel Sampling (several candidates in one form).\par
\medskip
\textbf{Revision forms}\par
F1 conditional rules; F2 worked examples; F3 reasoning procedure (reasoning steps or a workflow); F4 full rewrite.\par
\medskip
\textbf{Your task}\par
Choose a search strategy that addresses the current failures, and the revision form where the strategy takes one, taking previous evaluation results into account. Briefly explain how the feedback and history support your choice.
\end{minipage}%
\par\vspace{7pt}
\end{minipage}%
}
\endgroup
\caption{An abbreviated prompt template for adaptive selection of revision operators in \method. The prompt uses \emph{iteration} for what the text calls a round.}
\label{fig:strategy_prompt}
\end{figure}

%% file: sections/experimental_details.tex
\section{Experimental Details}
\label{app:experimental_details}

\subsection{Datasets and Execution Settings}
\label{app:dataset_settings}

Table~\ref{tab:dataset_splits} lists the fixed training and test sets used in Table~\ref{tab:main_results}. Any samples used for candidate generation, screening, validation, or final skill selection are drawn from the training set. Methods that require a fixed internal validation split construct it from this set.

\begin{table}[htbp]
\centering
\small
\caption{Dataset sizes and training budgets of target-LLM executions shared by all methods. One execution runs one skill on one training sample; a Mind2Web task counts once despite containing multiple action steps.}
\label{tab:dataset_splits}
\begin{tabular*}{\linewidth}{@{\extracolsep{\fill}}lrrr@{}}
\toprule
Dataset & Training & Test & Budget ($6|\train|$) \\
\midrule
DocVQA           & 500 & 1,000 & 3,000 \\
LiveMath         & 397 &   211 & 2,382 \\
Mind2Web         & 800 &   252 & 4,800 \\
SpreadsheetBench & 629 &   280 & 3,774 \\
\bottomrule
\end{tabular*}
\end{table}

\paragraph{DocVQA.}
We use a fixed subset of 500 questions from the official training split and 1,000 from the labeled validation split, which serves as our test set. Sampling approximately preserves question-type frequencies and excludes overlap in document identity between the two sets. The target receives a page image and a question. We score the extracted answer with ANLS, taking the maximum over reference answers.

\paragraph{LiveMath.}
We use LiveMathematicianBench releases from November 2025 through April 2026 for training and May through June 2026 for testing, with no overlap in source papers. Each question has five answer choices. The correct choice and four distractors are deterministically shuffled before being labeled A--E. The target receives the question and choices without a proof sketch; the score is one if the predicted label is correct and zero otherwise.

\paragraph{Mind2Web.}
We use a fixed set of 800 training tasks and 252 tasks from the official cross-task test split. Evaluation predicts each action from the annotated prior actions and candidate skill page elements. Element accuracy is computed within each task and then averaged over tasks. This measures element selection under the given action history, rather than end-to-end success in a live browser.

\paragraph{SpreadsheetBench.}
Training uses 629 tasks from the 912-task release after excluding the test tasks and empty instructions. Testing uses the 280-task split of SpreadsheetBench-Verified released with SkillOpt~\citep{yang2026skillopt}. The target generates one Python program from the instruction and workbook preview; the program is executed and the output workbook checked against the references. A task passes only if all its workbook cases pass. Training tasks usually have three cases, whereas test tasks have one.

\paragraph{Models and decoding.}
The first model setting uses \texttt{claude-haiku-4-5-20251001-v1:0} for target executions and \texttt{claude-sonnet-4-6} for optimization. The second uses \texttt{claude-opus-4-8} and \texttt{claude-opus-5}, respectively. Target output limits are 6,000 tokens for LiveMath and 16,384 for the other tasks. A Mind2Web task can require several target calls, one per action step. The initial skill states the task and required response format. For example, the DocVQA initial skill reads: ``You answer questions about document page images. You will be given a question and an image of a single document page. Answer the question using the content of the document image. Return the final answer inside \texttt{\textless answer\textgreater...\textless/answer\textgreater} tags.''

\subsection{Baseline Settings}
\label{app:baseline_settings}

\paragraph{GEPA.}
We use the GEPA optimizer~\citep{agrawal2026gepa} with a single editable system-prompt component, Pareto candidate selection, and reflection minibatches of three samples. The merge operation is disabled. GEPA's system-aware merge composes a child by choosing, for each module, the prompt of one parent, so with a single component it can only reproduce a parent. GEPA also reports the merge-free optimizer as its main method and GEPA+Merge as a separate variant whose gains it attributes to systems with several modules. The fixed training set is divided 80/20 for generation and internal validation.

\paragraph{TextGrad.}
We implement the textual-gradient update procedure of TextGrad~\citep{yuksekgonul2024textgrad} with minibatches of three samples. Each update is evaluated on the internal validation split and reverted if validation performance decreases. We use the same 80/20 partition rule as GEPA.

\paragraph{SkillOpt.}
We use the SkillOpt trainer~\citep{yang2026skillopt} with task adapters for the shared execution environments. Training batches contain 40 samples and reflection minibatches 8, and the number of edits merged into the skill per step decays from four to two along a cosine schedule. Validation gating, slow updates, and optimizer-side meta guidance are enabled.

\paragraph{Trace2Skill.}
We use the combined variant of Trace2Skill~\citep{ni2026trace2skill}, which extracts lessons from both successful and failed training trajectories and consolidates them into a skill. Three consolidation runs vary record order and grouping using seeds 41, 42, and 43. The final skill is selected by its score on the evolution samples, with a fallback to the initial skill if all candidate skills score lower. Task executions use the target LLM of the corresponding setting; analysis and consolidation use its optimizer LLM.

\paragraph{Iter-CoT.}
We follow the released pipeline of Iter-CoT~\citep{sun2024enhancing}, in which every construction call is made by the target LLM and no optimizer LLM is used. Each training sample is first answered with zero-shot chain of thought under the initial skill at temperature 0.7. Wrong answers are revised in the same conversation with the original revise prompt for up to six iterations, stopping early when an iteration corrects nothing, and the summary prompt then produces a complete reasoning process, kept only if its final answer is still correct. The final skill appends exemplars sampled once from this pool to the initial skill, eight on DocVQA and LiveMath and four on Mind2Web and SpreadsheetBench, without validation-based selection. On Mind2Web an exemplar is one action step. Revision calls count toward the training budget; summarization calls are logged but not counted.

\paragraph{Baseline candidate generation with the candidate evaluation of \method.}
The two indented rows of Table~\ref{tab:hybrid_generation} replace the candidate generation of \method with that of a baseline and keep everything else. In each round, the generator receives the current skill and its execution feedback on the round's batch, exactly the executions that \method already pays for, and proposes candidate skills that enter the unchanged candidate evaluation, saved candidate skills, and final skill selection under the same $6|\train|$ budget. The GEPA generator is GEPA's reflective prompt mutation applied to a minibatch of three samples from the batch~\citep{agrawal2026gepa}. GEPA's Pareto candidate pool and its minibatch-then-full evaluation are not used, because they perform the role that candidate evaluation performs here, and the parent is always the current skill of \method. The hybrid therefore isolates GEPA's proposal operator and does not evaluate GEPA's search as a whole. The SkillOpt generator is one SkillOpt step~\citep{yang2026skillopt}: its failure and success analysts propose edits, which are merged, ranked under the cosine edit schedule of Appendix~\ref{app:baseline_settings}, and applied to the current skill; its epoch-level slow update and its own selection gate are not used, since candidate evaluation takes the place of the latter. Both hybrids were run once on LiveMath and SpreadsheetBench.

\subsection{\method Settings}
\label{app:stratune_settings}
\paragraph{Batches and set sizes.}
Training samples are split into 16 stratified batches. Table~\ref{tab:stratune_settings} lists the screening quotas and validation set sizes. I3 submits at most two of its three candidates.

\begin{table}[htbp]
\centering
\small
\caption{Screening quotas of previously solved and randomly sampled samples, and further-validation set sizes (30\% of the training set), with Haiku/Sonnet.}
\label{tab:stratune_settings}
\begin{tabular*}{\linewidth}{@{\extracolsep{\fill}}lrrr@{}}
\toprule
& \multicolumn{2}{c}{Screening} & \\
\cmidrule(lr){2-3}
Dataset & Solved & Random & Further validation \\
\midrule
DocVQA           & 18 & 18 & 150 \\
LiveMath         & 24 & 24 & 119 \\
Mind2Web         & 18 & 18 & 240 \\
SpreadsheetBench & 18 & 18 & 189 \\
\bottomrule
\end{tabular*}
\end{table}

The floor $\epsilon_0$ of the screening gain is 0.001 on DocVQA and 0.0025 elsewhere and the validation gain $\epsilon_{\mathrm{v}}$ is 0.01 (Appendix~\ref{app:evaluation_rules}); a sample counts as solved at a score of 0.9 on continuous-score task, and 1.0 on the binary-score tasks. 

\subsection{Token Accounting of \method}
\label{app:token_accounting}
Table~\ref{tab:token_cost} breaks down the tokens that one complete \method training run consumes on each dataset with Haiku/Sonnet, counting the input and output tokens of every call and excluding test evaluation. Target tokens are spent by Haiku executing skills on training samples, for execution feedback, candidate evaluation, and final skill selection. Optimizer tokens are spent by Sonnet in three kinds of calls: strategy selection, the per-round choice of search strategy and revision form; candidate generation under I1, I2, and I3; and the preparation of execution feedback, which summarizes failures and successful trajectories for the generation calls.

Two observations follow. First, the optimizer LLM accounts for a small share of training tokens, between 1.1 and 22.7 percent of the target tokens; the largest share, on SpreadsheetBench, comes from feedback preparation. Second, strategy selection, the step that \method adds over methods with a fixed revision operator, uses between 6 and 11 percent of the optimizer tokens and at most 1.45 percent of the target tokens, the shares reported in Figure~\ref{fig:token_cost}.

\begin{table}[htbp]
\centering
\footnotesize
\setlength{\tabcolsep}{3.5pt}
\caption{Tokens consumed by one \method training run per dataset with Haiku/Sonnet. Target and optimizer tokens are in millions, strategy-selection tokens in thousands. Ratio is optimizer tokens as a percentage of target tokens.}
\label{tab:token_cost}
\begin{tabular}{lrrrrrrr}
\toprule
& \multicolumn{3}{c}{All optimizer calls} & \multicolumn{4}{c}{Strategy selection} \\
\cmidrule(lr){2-4}\cmidrule(lr){5-8}
Dataset & Target & Optimizer & Ratio (\%) & Calls & Tokens & Of optimizer (\%) & Of target (\%) \\
\midrule
DocVQA & 9.75 & 0.51 & 5.3 & 14 & 49 & 9.5 & 0.50 \\
LiveMath & 6.47 & 0.51 & 7.9 & 10 & 49 & 9.6 & 0.76 \\
Mind2Web & 45.00 & 0.50 & 1.1 & 16 & 57 & 11.3 & 0.13 \\
SpreadsheetBench & 7.20 & 1.63 & 22.7 & 25 & 104 & 6.4 & 1.45 \\
\bottomrule
\end{tabular}
\end{table}

%% file: sections/appendix_results.tex
\section{Additional Results}
\label{app:additional_results}

\subsection{Fixed Revision Operators on LiveMath}
\label{app:fixed_operators}
\textbf{Motivation.} The ablations in Table~\ref{tab:ablation_generation} fix one of the two choices, the search strategy or the revision form, and leave the other to the optimizer LLM. They therefore do not compare adaptive selection with a complete revision operator held fixed. This experiment closes that gap by fixing both choices at once.

\textbf{Setup.} We run \method on LiveMath with Haiku/Sonnet twelve times, each time allowing exactly one revision operator: one search strategy (I1, I2, or I3) paired with one revision form (F1, F2, F3, or F4). Candidate evaluation, saved candidate skills, final skill selection, and the training budget of $6|\train|$ are unchanged, so these runs differ from the adaptive run only in that the operator is fixed. Worked examples (F2) need solved training samples as material, so rounds without such samples are skipped for the F2 operators. Each operator is run once.

\begin{table}[htbp]
\centering
\small
\caption{LiveMath test scores of \method with one fixed revision operator (rows: search strategy, columns: revision form), compared with the initial skill and with adaptive selection of strategy and form (Table~\ref{tab:main_results}); Haiku/Sonnet, $6|\train|$ budget.}
\label{tab:fixed_operators}
\begin{tabular*}{\linewidth}{@{\extracolsep{\fill}}lrrrr@{}}
\toprule
& F1 rules & F2 examples & F3 procedure & F4 rewrite \\
\midrule
I1 direct revision      & 46.92 & 45.50 & 43.13 & 34.12 \\
I2 iterative refinement & 28.44 & 29.38 & 29.38 & 29.38 \\
I3 parallel sampling    & 37.91 & 36.49 & 63.51 & 58.29 \\
\midrule
Initial skill           & \multicolumn{4}{c}{29.38} \\
\method, adaptive strategy and form & \multicolumn{4}{c}{65.88} \\
\bottomrule
\end{tabular*}
\end{table}

\textbf{Results.} Table~\ref{tab:fixed_operators} reports the test scores. First, the best fixed operator is parallel sampling of reasoning procedures (I3 with F3), which scores 63.51, close to the 65.88 of adaptive selection. Second, every other operator falls well short: I3 with F4 by 7.6 points and the remaining ten by 19 to 37 points. The four I2 operators never change the initial skill, because iterative refinement restricted to one form produced no candidate skill that passed candidate evaluation. Third, adaptive selection reaches the level of the best fixed operator without knowing which operator that is. Identifying it here required all twelve runs, and Table~\ref{tab:ablation_generation} shows that the best fixed strategy differs across datasets, so the operator cannot be chosen in advance. The adaptive run did not itself use I3 with F3; it obtained its score through iterative refinement with full rewrites (I2 with F4), so more than one operator leads to a strong skill on LiveMath.

\subsection{Bandit Strategy Selectors on LiveMath and SpreadsheetBench}
\label{app:bandit_selectors}
\textbf{Motivation.} The ablations \emph{Rand.}, \emph{Rot.}, \emph{Once}, and \emph{Replay} in Table~\ref{tab:ablation_generation} remove the per-round decision, but none of them selects the strategy from the recorded outcomes by a rule. The \emph{Bandit} rows of Table~\ref{tab:ablation_generation} answer whether such a rule does as well as the optimizer LLM, which reads the failure types in the execution feedback in addition to the outcomes; this section gives the rules and the runs.

\textbf{Setup.} We replace the strategy choice of \method by three multi-armed bandit rules~\citep{kuleshov2014algorithms}, as in adaptive operator selection for evolutionary algorithms~\citep{fialho2010analyzing}, that see only the outcomes recorded in $H_t$: $\epsilon$-greedy on the mean reward of each strategy with $\epsilon=0.1$~\citep{sutton1998reinforcement}, UCB1~\citep{auer2002finite}, and Thompson sampling with a Beta prior~\citep{william1933likelihood,chapelle2011empirical}. The reward of a round is 1 if a candidate submitted by the chosen strategy replaced the current skill, 0.5 if it was saved for final skill selection, and 0 otherwise; untried strategies are chosen first. Everything else is unchanged: the same three strategies and four forms, the optimizer LLM still chooses the revision form, and candidate evaluation, saved candidate skills, final skill selection, and the $6|\train|$ budget are those of \method. We run each selector once on each dataset with Haiku/Sonnet, drawing the same training samples as the \emph{Replay} runs of Table~\ref{tab:ablation_generation} and the \emph{Rot.} runs on DocVQA, LiveMath, and Mind2Web, so these rows are directly comparable; the SpreadsheetBench \emph{Rot.} run with the same samples scores 45.71 (Table~\ref{tab:ablation_generation} reports a second run, 52.14).

\begin{table}[htbp]
\centering
\small
\caption{Test scores with the strategy chosen by the optimizer LLM (\method), by a fixed or replayed schedule, and by three bandit rules over the recorded outcomes; Haiku/Sonnet, $6|\train|$ budget. The schedule and bandit rows use the same training samples.}
\label{tab:bandit_selectors}
\begin{tabular*}{\linewidth}{@{\extracolsep{\fill}}lrrrr@{}}
\toprule
Strategy selection & DocVQA & LiveMath & Mind2Web & SpreadsheetBench \\
\midrule
\method, optimizer LLM (Table~\ref{tab:main_results}) & \textbf{92.12} & \textbf{65.88} & \textbf{47.49} & \textbf{56.43} \\
\emph{Rot.}, fixed schedule            & 91.74 & 65.40 & 46.06 & 45.71 \\
\emph{Replay}, copied sequence         & 92.00 & 51.66 & 46.18 & 53.21 \\
$\epsilon$-greedy                      & 91.94 & 55.45 & 46.09 & 49.64 \\
UCB1                                   & 90.92 & 38.39 & 46.02 & 53.21 \\
Thompson sampling                      & 91.41 & 45.02 & 46.07 & 53.57 \\
\midrule
Initial skill                          & 47.88 & 29.38 & 46.09 & 40.00 \\
\bottomrule
\end{tabular*}
\end{table}

\textbf{Results.} Table~\ref{tab:bandit_selectors} reports the test scores. All three bandit rules score below \method on all four datasets, by 0.2 to 1.2 points on DocVQA, 10.4 to 27.5 on LiveMath, 1.4 to 1.5 on Mind2Web, and 2.9 to 6.8 on SpreadsheetBench, and no bandit exceeds the best fixed or replayed schedule on any dataset. DocVQA does not separate the selectors: every bandit run accepts the same round-1 candidate as the other methods, a conditional rule from direct revision, and the runs end within 1.2 points of one another. On Mind2Web no bandit run accepts a candidate that survives to the end, and final skill selection returns the initial skill in all three runs, whereas \method accepts one update. The reason is visible in the runs. A run accepts zero to two candidate skills, so the reward a bandit receives is almost always zero and the rules fall back to their default behaviour. UCB1 keeps exploring and declares the three strategies almost uniformly (6, 6, and 5 rounds on LiveMath), and scores lowest on LiveMath. $\epsilon$-greedy commits to the first strategy that is rewarded, direct revision, which it declares in 18 of 22 rounds on SpreadsheetBench and in 27 of 30 rounds on Mind2Web, and scores lowest on SpreadsheetBench. Thompson sampling lies between the two. The optimizer LLM receives the same outcomes but also reads what the current skill gets wrong, and Table~\ref{tab:ablation_generation} shows that withholding the outcomes from it (\emph{No hist.}) also lowers the score. Both sources of information therefore matter, and a rule that sees only the outcomes does not replace the optimizer LLM at this budget. These are single runs, and LiveMath scores vary between runs of every configuration, so the LiveMath margins are indicative; the margins on the other three datasets are smaller but consistent across the three rules.

\subsection{Variability Across Optimization Runs}
\label{app:run_variability}
We assess the variability of \method using four independent runs with different seeds for each dataset--model setting. Table~\ref{tab:run_variability} reports the mean and standard deviation of the test scores. The standard deviations range from 0.40 to 1.16 points on the 0--100 scale. The main comparison in Table~\ref{tab:main_results} reports single-run results.

\begin{table}[htbp]
\centering
\small
\caption{Mean $\pm$ standard deviation of \method test scores (0--100) over four optimization runs per dataset--model setting.}
\label{tab:run_variability}
\begin{tabular*}{\linewidth}{@{\extracolsep{\fill}}lrrrr@{}}
\toprule
Setting & DocVQA & LiveMath & Mind2Web & SpreadsheetBench \\
\midrule
Haiku & $92.52 \pm 0.80$ & $65.40 \pm 1.16$ & $47.52 \pm 0.61$ & $56.16 \pm 0.90$ \\
Opus & $96.88 \pm 0.40$ & $81.16 \pm 0.80$ & $51.48 \pm 0.56$ & $72.41 \pm 0.70$ \\
\bottomrule
\end{tabular*}
\end{table}